\documentclass{article}
\usepackage{spconf,amsmath,graphicx}

\usepackage{times}  
\usepackage{helvet} 
\usepackage{courier}  
\usepackage[hyphens]{url}  
\usepackage{graphicx} 
\usepackage{caption} 

\usepackage{booktabs}
\usepackage{multirow}
\usepackage{float}
\usepackage{mathtools}
\usepackage{soul}
\usepackage{color}
\usepackage{enumitem}
\usepackage{hyperref}
\usepackage{diagbox}
\usepackage{makecell}
\usepackage[toc,page]{appendix}
\usepackage[colorinlistoftodos]{todonotes}
\usepackage{subfigure}
\usepackage{algorithm}
\usepackage{algpseudocode}
\usepackage{amsfonts}

\title{Multivariate Time Series Forecasting with Parallel Extraction of Long-Term Trends and Short-Term Fluctuations Framework}
\begin{document}
\name{Yifu Zhou$^1$, Ziheng Duan$^2$\sthanks{The first two authors contributed equally.},
 Haoyan Xu$^2$, Jie Feng$^2$,  Anni Ren$^2$,
 Yueyang Wang$^{3\dagger}$, Xiaoqian Wang$^4$\sthanks{Corresponding authors.}}

\address{$^1$The University of Queensland, Australia; $^2$Zhejiang University, China; \\ $^3$Chongqing University, China;
 $^3$Purdue University, USA}
%
\maketitle
\begin{abstract}
Multivariate time series (MTS) forecasting has been widely used in various fields. 
Reasonable prediction results can help people make decisions, avoid risks, and increase profits.
Generally, time series have two characteristics, namely long-term trends and short-term fluctuations. 
For example, stock prices will rise in the long term, but may fall slightly in the short term.
These two characteristics are often relatively independent of each other. 
Existing forecasting methods usually do not distinguish between the two characteristics and thus cannot fully extract inherent attributes of the time series. 
In this paper, we propose a framework that can capture the long-term trends and short-term fluctuations of time series in parallel in order to enhance MTS forecasting performance.
We formulate our model as a multi-task learning objective, the goal of which is to make predictions of long-term trends and short-term fluctuations as close as possible to the ground truth.
Experiments on three real-world datasets show that the proposed method uses more supervision information and can more accurately capture the changing trends of time series, thereby improving the prediction performance. 
\end{abstract}
\begin{keywords}
Multivariate Time Series Forecasting, Long-Term Trends, Short-Term Fluctuations
\end{keywords}
\section{Introduction}
\label{sec:intro}

Time series forecasting is a key challenge in many disciplines, including finance \cite{li2012intelligent}, industry \cite{blinowska1991non}, environment \cite{shen2013ensemble}, etc. 
Due to the diversity of the real world, a system usually contains multiple variables,
which makes the research of multivariate time series (MTS) forecasting particularly important.
\begin{figure}[t]
\centering
\includegraphics[width=0.6\linewidth]{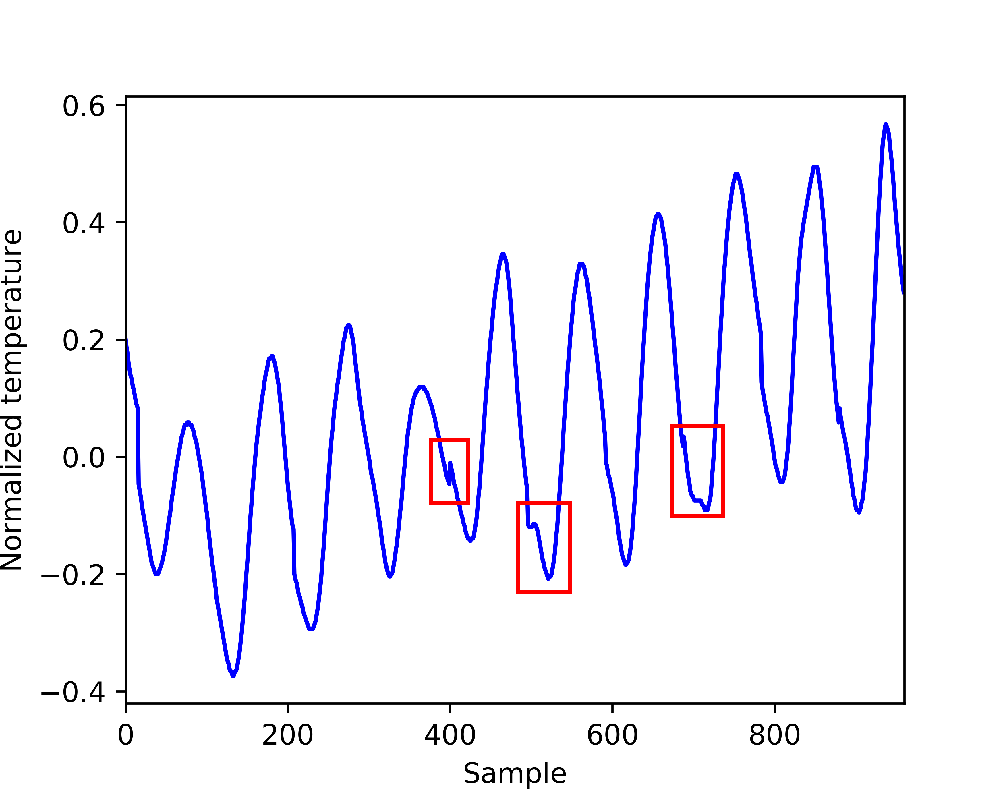}
\caption{Normalized temperature curve of a photovoltaic power plant. The blue solid line represents the standardized temperature, and the red boxes represent short-term fluctuations.}
\label{fig:1}
\vspace{-2mm}
\end{figure}

In most cases, time series collected from the real world have two characteristics, which are long-term trends and short-range fluctuations. 
As a concrete example in Fig. \ref{fig:1}, the temperature not only has periodic long-terms trends with the change of day and night, but also has fluctuations in the short term due to temporary weather reasons. 
Obviously, these two characteristics are relatively independent of each other, but both contain meaningful information that is important to the prediction of future values of the time series. 
In this case, the time series forecasting task can be decomposed into two parts, one is to predict its long-term trends, and the other part is to predict its short-term fluctuations. 
Together, the two constitute a complete forecasting result. 
Most existing time series forecasting methods take the original value of the time series as input to directly predict its future value. 
This confuses the two independent change rules and may reduce the prediction accuracy and interpretability of the model.

In economics, the value of the original time series is often used to represent the long-term trends, and the difference between time stamps is used as a measure of short-term fluctuations. 
Inspired by this representation, we design a framework to extract long-term trends and short-term fluctuations in parallel, named \emph{\underline{P}}arallel \emph{\underline{F}}orecasting \emph{\underline{Net}}work (PFNet). 
PFNet is composed of three sub-modules, which are long-term trends prediction module (LTPM), short-term fluctuations prediction module (SFPM), and information fusion module (IFM). 
Compared with direct forecasting, PFNet integrates more supervision information and can more accurately capture the different changing patterns of the time series, thereby improving the forecasting performance. Our main contributions are:
\vspace{-2mm}
\begin{itemize}[leftmargin=9pt]
\item We first propose an MTS forecasting framework to predict long-term trends and short-term fluctuations in parallel.
\item We construct a triplet loss function and integrate more supervision information to extract key features of time series.
\item We empirically validate that PFNet has better performance than state-of-the-art models on various benchmark datasets.
\end{itemize}

\begin{figure*}[t]
\centering
\includegraphics[width=0.9\linewidth]{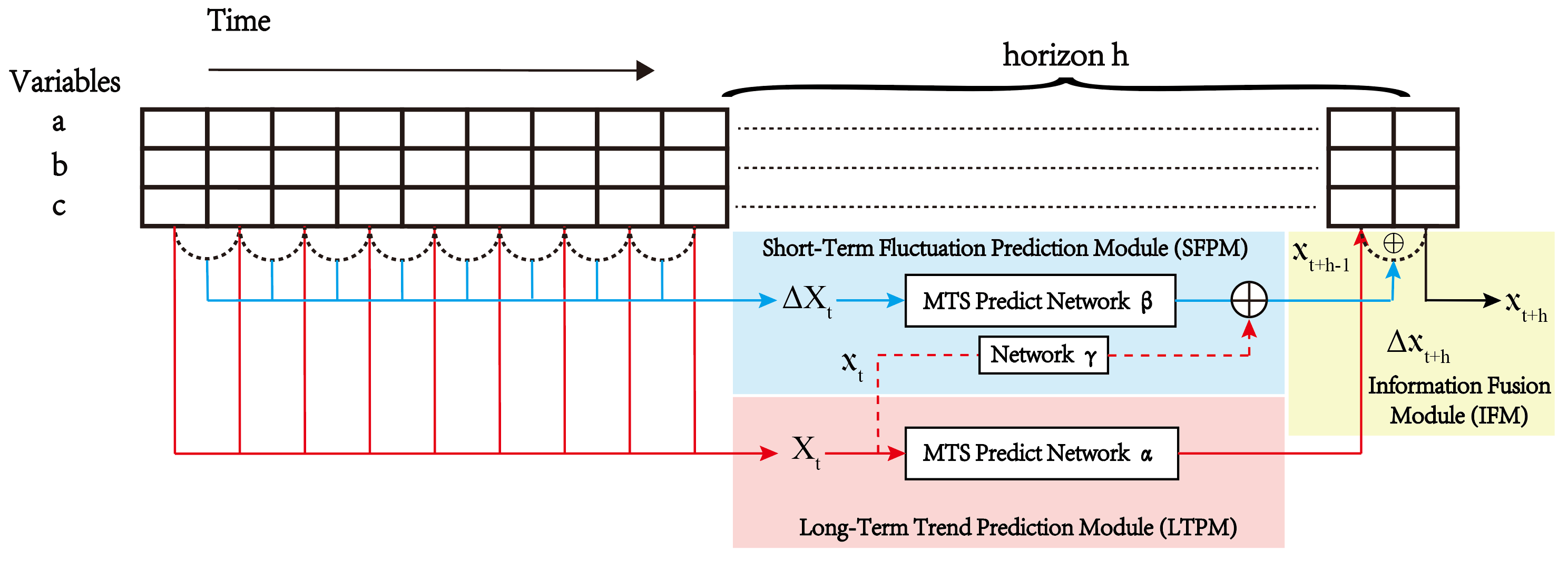}
\caption{The schematic of PFNet. The red line represents the data flow of the long-term trends of the time series, and the blue line represents the short-term fluctuations. We predict both of them in parallel through the LTPM and SFPM. Here we use Highway CNN as the network $\alpha$ and $\beta$, and MLP as the network $\gamma$. 
In this diagram $\Delta X_t \in \mathbb{R}^{N\times (t-1)}$ and $x_t \in \mathbb{R}^{N\times t}$. $x_t, x_{t+h-1}, x_{t+h}$ and $\Delta x_{t+h} \in \mathbb{R}^{N\times 1}$.
As shown by the red dotted line, part of the long-term trends ($x_t$) is used with $\Delta X_t$ to predict $\Delta x_{t+h}$. Finally, we add the predicted $x_{t+h-1}$ and $\Delta x_{t+h}$ to get the final result $x_{t+h}$.}
\label{fig:2}
\vspace{-2mm}
\end{figure*}
\vspace{-3mm}
\section{Framework}
\vspace{-2mm}
Given a matrix containing multiple observed MTS $X = \left[x_1, x_2,..., x_t\right]$, 
where $x_i \in \mathbb{R}^n$  and $n$ is the number of variables.
The main purpose of our forecasting model is to predict $x_{t+h}$, 
where $h$ is the \emph{horizon} ahead of the current time stamp.

%

\subsection{Vector Error Correction Model (VECM)}
As the most common used linear MTS forecasting method, VAR \cite{qiu2015robust} predicts $x_{t+h}$ in the following expression:
$x_{t+h} = \sum\limits_{i = 1}^{t} A_i^h x_i + \epsilon_t,$
where $A_i^h (i=1, 2, ...,t)$ are the regression coefficients when \textit{horizon} is $h$, and $\epsilon_t$ is the white noise vector. 
Based on this, VECM \cite{mukherjee1995dynamic} is proposed and according to it, $\Delta x_{t+h}$ can also be written as:
\vspace{-2mm}
\begin{equation}
\Delta x_{t+h} = \Pi x_t + \sum\limits_{k = 1}^{t} \Gamma_k \Delta x_{t-k} + \epsilon_t\,.
\label{equation:4}
\end{equation}
$\Pi$ and $\Gamma_k$ can 
be expressed by $A^1$, $A^2$, ..., $A^h$.
It can be seen that the short-term fluctuations to be predicted is related to two parts: the value of MTS at time $t$ and difference between previous time stamps. 
In our model, we replace the linear transformation in equation \ref{equation:4} with a neural network. 

\subsection{Long-Term Trends Prediction Module (LTPM)}
LTPM predicts long-term trends of MTS, which takes $X$ as input 
and predict $x_{t+h-1}$ instead of directly predicting $x_{t+h}$.
In this paper, we use Highway-CNN \cite{slimani2019compound} to deal with long-term trends prediction. 
Denote the nonlinear transformation of the multi-layer CNN as  $H(X,\theta)$, where $\theta$ represents model parameters. The output of Highway-CNN can be calculated using the following equation:
\vspace{-2mm}
\begin{equation}
y = H(X,\theta) \odot T(X,W_T) + W_LX \odot (1-T(X,W_T))\,,
\end{equation}
where $\odot$ is dot product and $W_L$ is a trainable linear mapping.
$W_T$ is a weight matrice. $T$ and $(1-T)$ represent the transform gate and carry gate respectively.
Here $T$ is calculated as:
$T(X,W_T) = 1/(1+e^{-W_TX}).$
Through LTPM, the predicted value $\hat x_{t+h-1}$ is obtained.

\subsection{Short-Term Fluctuations Prediction Module (SFPM)}
Refer to VECM, SFPM takes $\Delta X$ and $x_t$ as inputs to predict $\Delta x_{t+h}$, thereby characterizing its short-term fluctuations. Similar to LTPM, SFPM uses Highway-CNN to extract the features of $\Delta X$. In addition, a multilayer perceptron (MLP) structure is used for nonlinear transformation of $x_t$ . Therefore, the expression of the output vector $\hat x_{t+h}$ of SFPM is:
\vspace{-2mm}
\begin{equation}
\Delta \hat x_{t+h} = HighwayCNN(\Delta X)+MLP(x_t)\,.
\end{equation}

\subsection{Information Fusion Module (IFM)}
The final forecasting result is obtained by the superposition of long-term trends and short-term fluctuations. The IFM takes the output of LTPM and SFPM as input, and merges them to obtain the forecasting $\hat x_{t+h}$.
\vspace{-2mm}
\begin{equation}
\hat x_{t+h} = \hat x_{t+h-1} + \Delta \hat x_{t+h}\,.
\end{equation}

\subsection{Objective Function}
We use $L_1$ norm to construct a triple loss function (a weighted summation, using $c_1$ and $c_2$, of the $L_1$ loss for $x_{t+h}$, $x_{t+h-1}$ and $\Delta x_{t+h}$) and optimize the model via Adam algorithm.
\vspace{-2mm}
\begin{equation}
\begin{aligned}
     \min\limits_{\Theta}\sum\limits_{t}
     &\left\|\hat x_{t+h}-x_{t+h}\right\|_1 
     + c_1 \left\|\hat x_{t+h-1}-x_{t+h-1}\right\|_1 \\
     &+ c_2 \left\|\hat \Delta x_{t+h}-\Delta x_{t+h}\right\|_1\,.
\end{aligned}
\end{equation}

\begin{table*}[t]
\caption{MTS forecasting results under different horizons measured by RSE/RAE/CORR score over three datasets. Lower RSE and RAE and higher CORR values show better prediction performance.}
\centering
\scalebox{0.78}{
\begin{tabular}{lc|cccc|cccc|cccc}
\toprule
Dataset&&\multicolumn{4}{c|}{Exchange-Rate}& \multicolumn{4}{c|}{Energy} & \multicolumn{4}{c}{Nasdaq}\\
\midrule
&&\multicolumn{1}{c}{horizon}&\multicolumn{1}{c}{horizon}&\multicolumn{1}{c}{horizon}&\multicolumn{1}{c|}{horizon}

&\multicolumn{1}{c}{horizon}&\multicolumn{1}{c}{horizon}&\multicolumn{1}{c}{horizon}&\multicolumn{1}{c|}{horizon}

&\multicolumn{1}{c}{horizon}&\multicolumn{1}{c}{horizon}&\multicolumn{1}{c}{horizon}&\multicolumn{1}{c}{horizon}\\


Methods&Metrics
&\multicolumn{1}{c}{3} &\multicolumn{1}{c}{6} &\multicolumn{1}{c}{12} &\multicolumn{1}{c|}{24} 
&\multicolumn{1}{c}{3} &\multicolumn{1}{c}{6} &\multicolumn{1}{c}{12} &\multicolumn{1}{c|}{24} 
&\multicolumn{1}{c}{3} &\multicolumn{1}{c}{6} &\multicolumn{1}{c}{12} &\multicolumn{1}{c}{24} \\
\midrule

\multirow{3}{*}{\textsc{VAR}}
    &RSE& 0.0186 & 0.0262 & 0.0370 & 0.0505 & 0.1197 & 0.1314 & 0.1498 & 0.1791 & 0.0009 & 0.0015 & 0.0019 & 0.0042\\
    &RAE& 0.0141 & 0.0208 & 0.0299 & 0.0427 & 0.0448 & 0.0587 & 0.0793 & 0.1110 & 0.0008 & 0.0013 & 0.0017 & 0.0027 \\
    &CORR& 0.9674 & 0.9590 & 0.9407 & 0.9085 & 0.9394 & 0.8961 & 0.8259 & 0.6988 & 0.9959 & 0.9871 & 0.9818 & 0.8895 \\
\midrule

\multirow{3}{*}{\textsc{RNN}}
    &RSE& 0.0200	& 0.0262	& 0.0366	& 0.0527	& 0.1106	& 0.1162	& 0.1255	& 0.1322 & 0.0012	& 0.0015	& 0.0016	& 0.0021 \\
    &RAE& 0.0157	& 0.0209	& 0.0298	& 0.0442	& 0.0322	& 0.0422	& 0.0552	& 0.0680 & 0.0011	& 0.0014	& 0.0015	& 0.0020 \\
    &CORR& 0.9772	& 0.9688	& 0.9534	& 0.9272	& 0.9500	& 0.9141	& 0.8537	& 0.7636 & 0.9944	& 0.9925	& 0.9889	& 0.9817 \\
\midrule

\multirow{3}{*}{\textsc{MHA}}
    &RSE& 0.0194	& 0.0260	& 0.0360	& 0.0485	& 0.1103	& 0.1162	& 0.1260	& 0.1300 & 0.0010	& 0.0012	& 0.0016	& 0.0020 \\
    &RAE& 0.0151	& 0.0210	& 0.0298	& 0.0410	& 0.0335	& 0.0431	& 0.0561	& 0.0773	& 0.0009	& 0.0011	& 0.0014	& 0.0019 \\
    &CORR& 0.9779	& 0.9694	& 0.9547	& 0.9369	& 0.9474	& 0.9084	& 0.8422	& 0.7464	& 0.9961	& 0.9945	& 0.9908	& 0.9834 \\
\midrule

\multirow{3}{*}{\textsc{LSTNET}}
    &RSE& 0.0216	& 0.0277	& 0.0359	& 0.0482 & 0.1082	& 0.1160	& \textbf{0.1187}	& \textbf{0.1243}  & 0.0010	& 0.0013	& 0.0016	& 0.0021 \\
    &RAE& 0.0171	& 0.0226	& 0.0295	& 0.0404	&	0.0342	& 0.0430	& 0.0546	& 0.0688 & 0.0010	& 0.0012	& 0.0015	& 0.0019 \\
    &CORR& 0.9748	& 0.9679	& 0.9534	& 0.9353	&	0.9360	& 0.8984	& 0.8429	& 0.7714 & 0.9955	& 0.9933	& 0.9897	& 0.9829 \\
\midrule

\multirow{3}{*}{\textsc{MLCNN}}
    &RSE& 0.0172	& 0.0447	& 0.0519	& 0.0448	& 0.1113	& 0.1324	& 0.1225	& 0.1331	& 0.0009	& 0.0011	& 0.0015	& 0.0020\\
    &RAE& 0.0129	& 0.0334	& 0.0422	& 0.0378	& 0.0345	& 0.0501	& 0.0570	& 0.0664	& 0.0008	& 0.0010	& 0.0013	& 0.0018\\
    &CORR& 0.9780	& 0.9610	& 0.9550	& 0.9406	& 0.9481	& 0.9065	& 0.8424	& 0.7685	& 0.9974	& 0.9956	& 0.9920	& 0.9839\\
\midrule

\multirow{3}{*}{\textsc{MTGNN}}
    &RSE& 0.0194	& 0.0253	& 0.0345	& 0.0447	&	0.1127	& \textbf{0.1149}	& 0.1203	& 0.1273	& 0.0011	& 0.0012	& 0.0042	& 0.0080 \\
    &RAE& 0.0156	& 0.0206	& 0.0283	& 0.0376	& 0.0302	& 0.0463	& 0.0541	& \textbf{0.0647}	& 0.0011	& 0.0012	& 0.0043	& 0.0070 \\
    &CORR& 0.9782	& 0.9711	& 0.9564	& 0.9370	& 0.9505	& 0.9048	& \textbf{0.8565}	& \textbf{0.7745}	& 0.9966	& 0.9944	& 0.9843	& 0.9814 \\
\midrule

\multirow{3}{*}{\textsc{PFNet}}
    &RSE& \textbf{0.0156}	& \textbf{0.0229}	& \textbf{0.0332}	& \textbf{0.0437}	&	\textbf{0.1074} & 0.1159 &	0.1221 & 0.1312	&	\textbf{0.0008}	& \textbf{0.0009}	& \textbf{0.0013}	& \textbf{0.0020}\\
    &RAE& \textbf{0.0121}	& \textbf{0.0180}	& \textbf{0.0268}	& \textbf{0.0367}	&	\textbf{0.0278}	& \textbf{0.0381}	&\textbf{0.0524}	& 0.0714	&	\textbf{0.0007}	& \textbf{0.0009}	& \textbf{0.0013}	& \textbf{0.0019}\\
    &CORR& \textbf{0.9813}	& \textbf{0.9732}	& \textbf{0.9583} & \textbf{0.9386} 	&	\textbf{0.9610}	& \textbf{0.9226}	& 0.8562	& 0.7600	&	\textbf{0.9987}	& \textbf{0.9968}	& \textbf{0.9931}& \textbf{0.9854}\\
\midrule

\end{tabular}}
\label{sec:table1}
\vspace{-1mm}
\end{table*}

\vspace{-3mm}
\section{Experiments}
\vspace{-2mm}

\subsection{Experimental Setup}

We use three benchmark datasets which are publicly available:
\textbf{Exchange-Rate}\footnote{https://github.com/laiguokun/multivariate-time-series-data/tree/master/exchange\_rate} is the exchange rates of eight foreign countries collected from 1990 to 2016, collected per day.
\textbf{Energy} \cite{candanedo2017data} measurements of 26 different quantities related to appliances energy consumption in a single house for 4.5 months, collected per 10 minutes.
\textbf{Nasdaq} \cite{qin2017dual} represents the stock prices which are selected as the multivariable time series for 82 corporations, collected per minute.

We apply three conventional evaluation metrics to evaluate the performance in multivariate time series prediction: Relative Squared Error (RSE), Relative Absolute Error (RAE) and Empirical Correlation Coefficient (CORR).
The comparison methods include:
\textbf{VAR} \cite{hamilton1994time} stands for the well-known vector regression model.
\textbf{RNN} \cite{chung2014empirical} is the Recurrent Neural Network using GRU cell with AR components. 
\textbf{MHA} \cite{NIPS2017_7181}, or MultiHead Attention, stands for multihead attention components in the Transformer model.
\textbf{LSTNet} \cite{DBLP:journals/corr/LaiCYL17} shows great performance by modeling long-term and short-term temporal patterns of MTS data.
\textbf{MLCNN} \cite{cheng2019towards} is a novel multi-task deep learning framework which adopts the idea of fusing foreacasting information of different future time.
\textbf{MTGNN} \cite{wu2020connecting} is a joint framework for modeling multivariate time series data generally from a graph-based perspective with graph neural networks.
\textbf{PFNet} is our proposed Parallel Forecasting Network, which predicts long-term trends and short-term fluctuations of time series in parallel.


For the training details, we conduct grid search on tunable hyper-parameters on each method over all datasets. 
Specifically, the same grid search range of input window size for each method is set from \{$2^0$, $2^1$, ..., $2^9$\} and different hyper-parameters are chosen for each method to achieve their best performance on this task. For RNN-GRU and LSTNet, the hidden dimension of Recurrent and Convolutional layer is chosen from \{10, 20, ..., 100\}. 
For LSTNet, the skip-length $p$ is chosen from \{0, 12, .., 48\}. 
For MLCNN, the hidden dimension of recurrent and convolutional layer is chosen from \{10, 25, 50, 100\}. 
For PFNet, the kernal size of CNN is 3 and the size of highway window of CNN component is chosen from \{4, 8, 16, 32\}. During the training phase, the batch size is 128 and the learning rate is 0.001. 
The Adam algorithm is used to optimize the parameters of our model.

\subsection{Main Results}
Table \ref{sec:table1} shows the evaluation results.
Following the test settings of \cite{DBLP:journals/corr/LaiCYL17}, we use each model for time series predicting on future moment $\{t+3, t+6, t+12, t+24\}$.
The best results for each metrics on each dataset are set bold in Table \ref{sec:table1}. 
We save the model that has the best performance on validation set based on RSE or RAE metric after training 1000 epochs for each method and use this model to test and record the results. 

Overall, the performance of VAR is weaker than other baselines, which shows that the linear regression method is not conducive to fully extracting the variety of information in the time series. 
MTGNN or TEGNN achieves the best in some cases,
which shows the superiority of using graph structure to exploit the relationships among different variables in multivariate time series.
The results in the Table \ref{sec:table1} indicate that our proposed PFNet outperforms these baselines in most cases. 

It is worth noting that, as models for predicting the value of multiple time points in the future, PFNet performs better than MLCNN. 
The reason could be that LTPM and SFPM can better capture the long-term and short-term characteristics of the time series, and a single model may be easier to confuse both. 
PFNet uses the idea of parallel forecasting to capture the long-term trends and short-term fluctuations of multivariate time series. Thus it can break through the restriction that traditional methods and other deep learning methods cannot use both of them. 

\begin{figure}[t]
\centering    
	\centering          
	\includegraphics[scale=0.12]{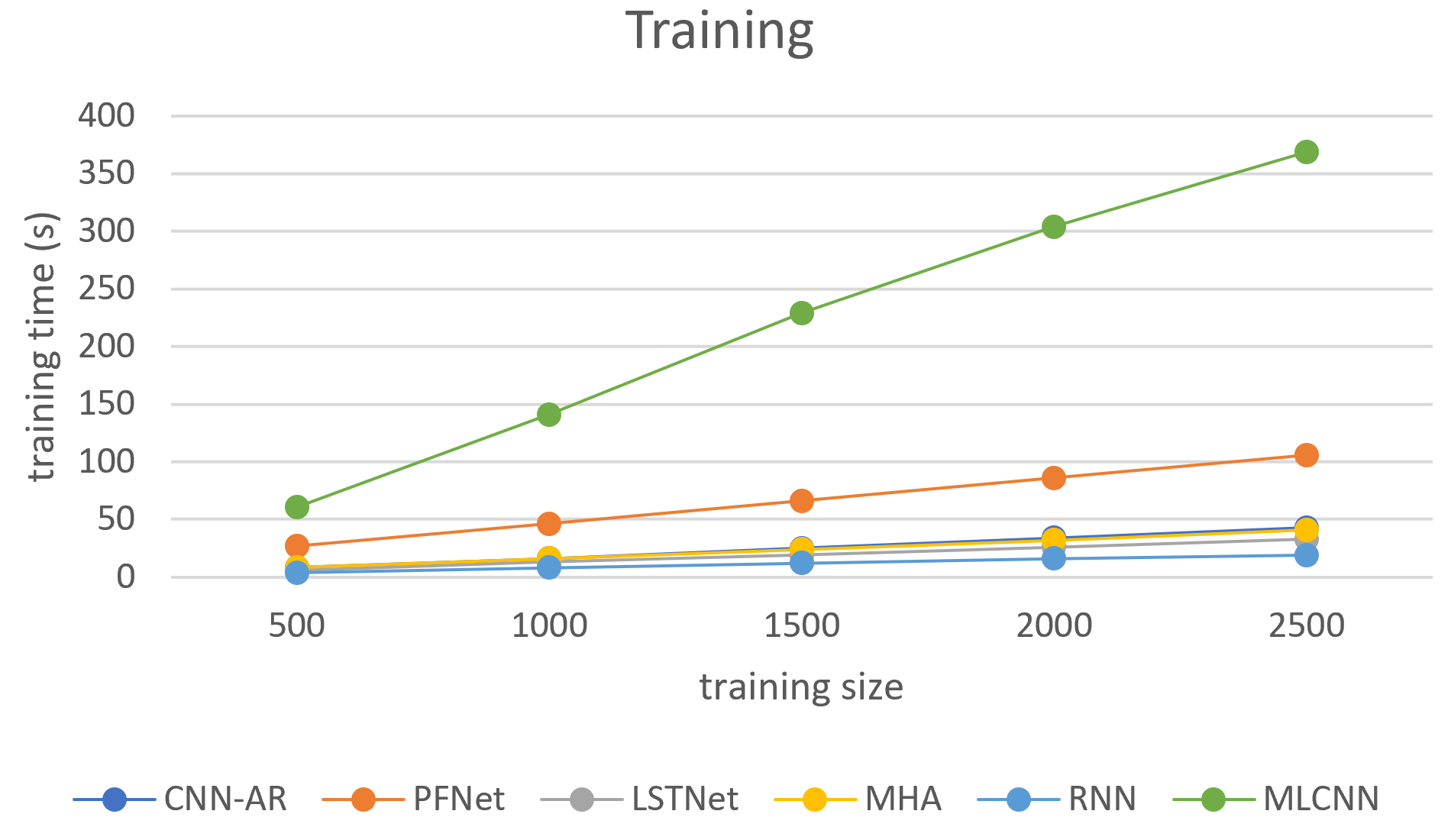}   
\caption{Result of time complexity comparison.} 
\label{fig:3}  
\vspace{-3mm}
\end{figure}

\vspace{-3mm}
\subsection{Time Complexity Analysis}

Although the proposed PFNet architecture designs three modules, it does not require much time. 
Following \cite{cheng2020towards}, we compare the performance of all models as a function of sample size and display the results on the NASDAQ dataset in Figure \ref{fig:3} to prove PFNet's efficiency. 
The training and testing time of our PFNet model is about twice as long as that of other simple baselines. 
This is in line with our expectations.
Because we not only forecast the long-term trends but also forecast the short-term fluctuations and the time used for each part is similar to that of an ordinary prediction network. 
It is worth noting that when processing high-dimensional time series, the time required for training and prediction of MLCNN, which also adopts the idea of multi-task learning, costs 3.6 and 6.5 times compared with our model PFNet. 
This proves the efficiency of our multi-task learning framework: there is no complicated network structure, and good performance can be achieved by combining this parallel forecasting framework with ordinary prediction networks (like highway-CNN in this paper).

\begin{table}[t]
\caption{Ablation Study.}
\centering
\scalebox{0.78}{
\begin{tabular}{lc|cccc} 
\toprule
Dataset&&\multicolumn{4}{c}{Exchange-Rate}\\
\midrule
&&\multicolumn{1}{c}{horizon}&\multicolumn{1}{c}{horizon}&\multicolumn{1}{c}{horizon}&\multicolumn{1}{c}{horizon}\\

Methods&Metrics
&\multicolumn{1}{c}{3} &\multicolumn{1}{c}{6} &\multicolumn{1}{c}{12} &\multicolumn{1}{c}{24} \\
\midrule

\multirow{3}{*}{\textsc{LTPM}}
    &RSE& 0.0194	& 0.0250	& 0.0342	& 0.0466	\\
    &RAE& 0.0150	& 0.0201	& 0.0281	& 0.0392	\\
    &CORR& 0.9776	& 0.9694	& 0.9548	& 0.9360	\\
\midrule

\multirow{3}{*}{\textsc{PFNet-xt}}
    &RSE& 0.0168	& 0.0234	& 0.0336	& 0.0449	\\
    &RAE& 0.0129	& 0.0187	& 0.0274	& 0.0374	\\
    &CORR& 0.9744	& 0.9723	& 0.9570	& 0.9374	\\
\midrule

\multirow{3}{*}{\textsc{PFNet-RNN}}
    &RSE& 0.0159	& 0.0239	& 0.0355	& 0.0557	\\
    &RAE& 0.0121	& 0.0191	& 0.0288	& 0.0165	\\
    &CORR& 0.9806	& 0.9719	& 0.9567 & 0.9359	\\
\midrule

\multirow{3}{*}{\textsc{PFNet-LSTNet}}
    &RSE& 0.0181	& 0.0232	& 0.0330	& 0.0484	\\
    &RAE& 0.0144	& 0.0184	& 0.0270	& 0.0410	\\
    &CORR& 0.9797	& 0.9726	& 0.9572	& 0.9348	\\
\midrule

\multirow{3}{*}{\textsc{PFNet}}
    &RSE& \textbf{0.0156}	& \textbf{0.0229}	& \textbf{0.0332}	& \textbf{0.0437}	\\
    &RAE& \textbf{0.0121}	& \textbf{0.0180}	& \textbf{0.0268}	& \textbf{0.0367}	\\
    &CORR& \textbf{0.9813}	& \textbf{0.9732}	& \textbf{0.9583} & \textbf{0.9386} 	\\
\midrule
\end{tabular}}
\label{sec:table2}
\vspace{-5mm}
\end{table}

\subsection{Ablation Study}
We conduct an ablation study on exchange-rate dataset to validate the effectiveness of key components of PFNet.
We name PFNet without different components as follows:

\textbf{LTPM}: PFNet without short-term fluctuations prediction module. 
Here we use multi-layer CNN with AR components \cite{lecun1995convolutional} to perform MTS forecasting tasks.
\textbf{PFNet-RNN}: PFNet that replaces CNN-AR module with RNN.
\textbf{PFNet-LSTNet}: PFNet that replaces CNN-AR module with LSTNet.
\textbf{PFNet-xt}: PFNet without MLP in short-term fluctuations prediction module. That means we only use $\Delta X$ to predict $\Delta \hat x_{t+h}$.

Table \ref{sec:table2} shows the comparison results. The important conclusions of these results are as follows: 
(1) \textbf{PFNet} has the best performance, compared with these variants. 
(2) After removing the short-term fluctuations prediction module, as shown by \textbf{LTPM}, the performance will drop, which proves that short-term fluctuations are useful for time series forecasting and PFNet takes advantage of this. 
(3) The relatively weak performance after removing MLP in SFPM (\textbf{PFNet-xt}) also proves the contribution of introducing $x_t$ in forecasting short-term fluctuations.
(4) As shown in \textbf{PFNet-RNN} and \textbf{PFNet-LSTNet}, even if the CNN-AR module is replaced by RNN or LSTNet, the results are also competitive, which proves that the framework of parallel forecasting is robust. 

\vspace{-2mm}
\section{Conclusion}
\vspace{-2mm}
In this paper, we propose a MTS forecasting framework, PFNet,
to capture the long-term trends and short-term fluctuations of time series in parallel. 
Inspired by the idea of multi-task learning, PFNet uses the original value and its difference between time stamps to optimize MTS forecasting problems.
Experiments on three real-world datasets show that our model outperforms 6 baselines in terms of the three metrics (RAE, RSE and CORR). 
For future research, different models can be incorporated into our framework
to improve prediction accuracy and make PFNet more robust.

\bibliographystyle{IEEEbib}
\bibliography{refs}

\end{document}